\documentclass{article}
\usepackage[final]{neurips_2026}
\usepackage[utf8]{inputenc}
\usepackage[T1]{fontenc}
\usepackage{booktabs}
\usepackage{hyperref}
\usepackage{url}
\usepackage{microtype}
\microtypesetup{expansion=false}

\title{The Evidence Ladder for Reinforcement Learning in Healthcare:\\
From Retrospective Policies to Trusted Interventions}
\author{Yunfan Zhao}

\begin{document}
\maketitle

\begin{abstract}
Reinforcement learning (RL) offers a natural language for healthcare decisions
whose consequences unfold over time, yet most reported progress remains far
from routine intervention. Existing surveys organize the field by algorithm or
clinical application. We instead review healthcare RL through an
\emph{evidence ladder}: problem formulation, retrospective identification,
policy estimation, stress testing, prospective evaluation, and lifecycle
monitoring. This view connects clinical treatment, patient engagement, and
health-system operations while exposing a recurring gap: evidence that a policy
scores well in a historical dataset is not evidence that it will improve care.
We synthesize the assumptions and failure modes at each rung, identify what
evidence can and cannot transfer across settings, and propose reporting
practices for cumulative evaluation. Restless bandits are included as one
special case, not as the organizing framework. The central lesson is that
healthcare RL should be evaluated as an intervention embedded in a changing
sociotechnical system, rather than only as an optimizer of a retrospective
reward.
\end{abstract}

\section{Introduction}

Healthcare is sequential. Clinicians adjust treatment as physiology changes,
digital systems adapt messages to patient response, and health systems allocate
staff and attention under evolving demand. RL formalizes such problems as
repeated decisions with delayed consequences \citep{sutton2018reinforcement}.
Applications now span sepsis and intensive care, mechanical ventilation,
chronic disease, mobile health, and resource allocation
\citep{komorowski2018artificial,raghu2017continuous,peine2021development,
nahum2018microrandomized,boehmer2025optimizing}.

The field has already been surveyed from several valuable perspectives. Broad
reviews categorize healthcare applications and RL algorithms
\citep{yu2021reinforcement}; critical-care reviews examine clinical decision
support \citep{liu2020reinforcement}; methodological guidance emphasizes risks
in observational cohorts \citep{gottesman2019guidelines}; and a recent primer
introduces RL to clinical audiences \citep{jayaraman2024primer}. A newer survey
frames RL as a transition from prediction toward agentive healthcare AI
\citep{perera2025beyond}. Repeating those taxonomies would add little.

This survey asks a different question: \emph{what evidence is required to move
from a learned policy to a trustworthy healthcare intervention?} We organize
the literature as an evidence ladder rather than by disease or algorithm. The
ladder makes comparisons possible across three families of decisions: (i)
clinical treatment, where actions directly alter care; (ii) patient-facing
engagement, where actions influence adherence or participation; and (iii)
health-system operations, where actions allocate scarce capacity. These
families differ in risk, feedback, and feasible experimentation, but face a
shared progression from formalization to deployment.

\paragraph{Scope and review method.}
We conducted a targeted narrative review centered on prior healthcare-RL
surveys and guidelines, then followed their foundational and representative
application references through 2025. We prioritized peer-reviewed work on
clinical policy learning, offline evaluation, adaptive interventions, safety,
fairness, and deployment. We also included general RL methods when they define
an evidentiary requirement used in healthcare. This is not a meta-analysis:
heterogeneous outcomes and scarce prospective evaluations make pooled effect
estimates inappropriate. Its contribution is a cross-setting framework for
interpreting evidence, not an exhaustive catalog.

\section{A Map of Healthcare RL}

An RL formulation specifies states, actions, transitions, rewards, horizons,
and a policy. These objects have different meanings across healthcare settings.
In treatment, the state is an imperfect record of patient condition, actions
are clinical choices, and rewards encode multiple short- and long-term outcomes.
In engagement, actions may be messages or recommendations and feedback may be
sparse, behavioral, and nonstationary \citep{seow2024improving}. In operations,
actions distribute staff, monitoring, appointments, or outreach across many
people, coupling individual outcomes through shared constraints
\citep{behari2024decision,boehmer2025optimizing}.

This map cuts across algorithmic categories. Dynamic treatment regimes and
Q-learning predate much of modern deep RL \citep{murphy2003optimal,
chakraborty2014dynamic}. Offline RL learns from fixed records; online RL adapts
through interaction; contextual bandits optimize shorter-horizon choices; and
constrained or safe RL represents limits on undesirable outcomes. Restless
bandits are useful for repeated allocation, but they are only one point in this
larger design space. Generalizable and communication-aware bandit policies
illustrate advances in scalable allocation \citep{zhao2024pretrained,
zhao2024zero,zhao2025bandit}, while implicit policy representations illustrate
that large action spaces can also be addressed outside bandit models
\citep{zhao2022implicit}.

\subsection{Clinical treatment and physiologic control}

Acute-care studies form the most visible branch of healthcare RL. Sepsis has
served as a recurring benchmark because treatment is sequential, outcomes are
consequential, and large observational cohorts are available. The AI Clinician
and related studies learn fluid and vasopressor strategies from intensive-care
records \citep{komorowski2018artificial,raghu2017continuous,rashidian2018deep,
peng2018improving}. Other work studies mechanical ventilation, sedation, and
weaning \citep{prasad2017reinforcement,peine2021development,yu2019reinforcement},
anesthesia control \citep{moore2014reinforcement,padmanabhan2015closed}, and
laboratory-test selection \citep{cheng2019dynamically}. These applications
illustrate both RL's appeal and its core identification problem: the sickest
patients receive different treatments for reasons incompletely captured in the
record.

Longer-horizon treatment planning broadens the challenge. Dynamic treatment
regime methods formalize sequences of treatment rules and provide statistical
foundations for Q-learning and actor--critic estimation
\citep{robins2004optimal,murphy2005generalization,zhao2015doubly,
schulte2014qlearning}. Applications include cancer treatment and radiotherapy
\citep{zhao2009reinforcement,tseng2017deep,jalalimanesh2017simulation}, epilepsy
and neurostimulation \citep{guez2008adaptive}, and glycemic control
\citep{teixeira2018reinforcement,fox2020reinforcement}. These studies differ
substantially in action authority: some generate hypotheses, some support a
clinician, and some control a device. The required evidence should rise with
the immediacy and irreversibility of the action.

\subsection{Chronic care and adaptive engagement}

Chronic disease management often combines clinical decisions with repeated
behavioral support. Mobile and wearable systems can adapt notifications,
coaching, activity suggestions, or adherence support at a much faster cadence
than clinic visits. The HeartSteps line of work helped establish
micro-randomized trials and contextual-bandit methods for just-in-time adaptive
interventions \citep{korinek2018adaptive,liao2020personalized,
nahum2018microrandomized,dempsey2015microrandomized}. Related research studies
physical activity, stress, oral-health behavior, and medication adherence
\citep{rabbi2015moodrhythm,tewari2017ads,forman2019first,
trella2022designing}. Here reward is frequently a short-term behavioral proxy;
whether optimizing it improves durable health remains a separate question.

Engagement also creates strategic feedback. Repeated messaging can cause
habituation or burden, and missing responses can reflect disengagement rather
than a neutral outcome. Cognitive models and recommendation research can help
separate response propensity from intervention effect
\citep{seow2024improving,liao2016personalized}. Bandit and RL methods are useful
when rapid experimentation is ethical, but randomization probabilities,
availability conditions, and burden constraints should be part of the reported
policy rather than hidden implementation details.

\subsection{Operations, allocation, and population health}

Healthcare decisions also occur above the individual-treatment level.
Sequential optimization has been applied to appointment scheduling, bed and
operating-room management, ambulance positioning, screening, and outreach
\citep{arnold2008patient,patrick2008dynamic,maxwell2010ambulance,
zhang2015dynamic,deo2020operational}. These problems couple people through
capacity: giving an appointment, monitor, or outreach call to one person can
delay service for another. Consequently, policy value must include congestion,
waiting, continuity, and distributional effects, not only the predicted benefit
of the selected individual.

Public-health allocation illustrates this coupling. Adaptive outreach and
monitoring systems use limited interventions across many beneficiaries
\citep{mate2019collapsing,killian2019restless,behari2024decision,
boehmer2025optimizing}. Restless bandits can exploit decomposable dynamics, but
other formulations---constrained MDPs, approximate dynamic programs, queues,
and simulation optimization---may better capture shared resources or network
effects \citep{altman1999constrained,powell2011approximate}. The model should be
chosen for the institution's decision structure, not because a particular RL
benchmark is convenient.

\section{Methodological Foundations}

\subsection{Offline policy learning}

Healthcare RL is usually an offline-learning problem. General treatments of
batch RL emphasize distributional shift between the behavior and learned
policies \citep{lange2012batch,levine2020offline}. Deep methods address this
shift through behavior regularization, pessimism, or conservative value
estimation \citep{fujimoto2019offpolicy,wu2019behavior,kumar2020conservative,
kostrikov2022offline}. Model-based approaches learn transitions and plan within
an estimated environment \citep{janner2019trust,dietterich2000hierarchical},
while uncertainty-aware methods restrict decisions unsupported by data
\citep{laroche2019safe,kidambi2020morel}. None eliminates the need to ask why
the historical action was selected and which relevant variables were absent.

Representation learning introduces another layer. Recurrent and latent-state
models can summarize irregular longitudinal records, but a compact predictive
state need not preserve treatment effect modifiers
\citep{hausknecht2015deep,killian2020empirical}. Missingness itself may encode
clinical attention; imputation can erase that signal or create artificial
trajectories \citep{lipton2016learning,che2018recurrent}. Pretrained policies and
generalizable representations may improve sample efficiency
\citep{zhao2024pretrained,zhao2024zero}, yet transfer claims require new-site
coverage and calibration evidence. More general advances in scalable neural
kernels and geometric representation learning may also support efficient model
adaptation and structured clinical data processing
\citep{sehanobish2024scalable,choromanski2023efficient}.

\subsection{Causal reasoning and off-policy evaluation}

Sequential causal inference and RL share vocabulary but answer different
questions. Potential-outcome and structural-causal frameworks clarify
consistency, positivity, and sequential exchangeability
\citep{robins1986new,hernan2020causal,pearl2009causality}. Marginal structural
models, g-computation, and doubly robust estimators provide tools for treatment
effects under time-varying confounding \citep{robins2000marginal,bang2005doubly,
vanderlaan2011targeted}. RL contributes policy search and long-horizon value
estimation. Credible healthcare work should state where causal identification
enters and avoid treating an MDP assumption as proof that confounding has been
resolved.

OPE is the bridge between a candidate policy and logged data. Importance
sampling is unbiased under strong support but can have extreme variance;
direct modeling is stable but sensitive to misspecification; doubly robust and
weighted estimators combine elements of both
\citep{precup2000eligibility,jiang2016doubly,thomas2016data,
farajtabar2018more}. More recent work studies marginalized ratios and efficient
inference for long-horizon settings \citep{liu2018breaking,kallus2020double,
uehara2020minimax}. In healthcare, estimator agreement should be treated as a
diagnostic, not as permission to ignore their shared assumptions.

\subsection{Safety, constraints, and human authority}

Safe RL distinguishes uncertainty about value from constraints on unacceptable
behavior. General surveys describe risk-sensitive objectives, constrained
optimization, and shielding \citep{garcia2015comprehensive,amodei2016concrete,
dulac2021challenges}. Constrained MDP methods optimize return subject to expected
cost limits \citep{achiam2017constrained,chow2018lyapunov}, while robust and
distributionally robust approaches seek protection against model error
\citep{iyengar2005robust,nilim2005robust,delage2010distributionally}. For
clinical use, expected constraints may be insufficient when a rare violation
is severe; action-level guardrails and clinician authority may be necessary.

Human involvement changes the policy being evaluated. Clinicians may accept,
modify, delay, or ignore recommendations, creating a joint human--AI policy.
Automation bias and alert fatigue can produce harms even when isolated model
accuracy is high \citep{parasuraman2010complacency,ancker2017effects,
cabitza2017unintended}. Evaluation should therefore record presentation,
override mechanisms, response time, and organizational escalation. Explanations
may support review, but post-hoc plausibility is not evidence of causal
faithfulness \citep{rudin2019stop,ghassemi2021false}.

\subsection{Fairness and distributional value}

Healthcare inequities can enter through access, measurement, labels, historical
treatment, and deployment. Widely used risk scores have demonstrated how cost
proxies can encode racial disparities \citep{obermeyer2019dissecting}; general
frameworks distinguish measurement, representation, and aggregation harms
\citep{suresh2021framework,selbst2019fairness}. Sequential systems add feedback:
allocating fewer resources can reduce observed benefit and justify future
under-allocation. Fair RL and long-term fairness research begins to address
these dynamics \citep{jabbari2017fairness,wen2021algorithms,verma2024group}.

No universal fairness metric resolves a healthcare allocation dispute.
Equality of action, opportunity, benefit, and health outcome imply different
policies, especially under scarcity. Studies should report multiple
distributional outcomes and make the normative choice explicit
\citep{rajkomar2018ensuring,char2018implementing}. Subgroup estimates also need
uncertainty: a noisy parity result is not evidence of equal effect.

\begin{table}[t]
\caption{The evidence ladder. Higher rungs depend on, but do not replace, lower
rungs.}
\label{tab:ladder}
\centering
\small
\begin{tabular}{p{0.16\linewidth}p{0.32\linewidth}p{0.38\linewidth}}
\toprule
Rung & Central question & Typical evidence \\
\midrule
Formulation & Is the decision problem clinically faithful? & Stakeholder review;
state/action/reward justification \\
Identification & Can historical data support the comparison? & Coverage,
confounding, missingness, temporal audits \\
Estimation & Does the candidate improve expected outcomes? & Multiple off-policy
estimators, uncertainty, baselines \\
Stress testing & Does performance survive plausible change? & Subgroups,
sensitivity, shift, worst-case tests \\
Prospective study & Does the intervention help in workflow? & Silent trials,
micro-randomization, controlled rollout \\
Lifecycle & Does value persist after deployment? & Monitoring, drift response,
re-audit, governance \\
\bottomrule
\end{tabular}
\end{table}

\section{The Evidence Ladder}

\paragraph{1. Formulation: establish clinical meaning.}
The first failure may occur before learning begins. Electronic health records
are partial observations shaped by clinician behavior; an ``action'' may record
an order rather than its delivery; and a convenient outcome may be a poor proxy
for patient benefit. Reward design is especially consequential because a policy
can exploit omissions while remaining mathematically optimal. Guidelines
therefore recommend clinical justification of states, actions, time steps, and
outcomes before comparing algorithms \citep{gottesman2019guidelines,
futoma2020cautionary}. Predict-then-optimize research similarly shows that
predictive fit and decision quality need not coincide
\citep{elmachtoub2023estimate}. Formulation evidence should document who chose
the objective, whose outcomes count, and which safety rules are inviolable.

\paragraph{2. Identification: ask what the data can reveal.}
Most healthcare RL is offline because unsafe exploration is unacceptable.
Historical trajectories, however, were generated by clinicians and institutions
whose choices depend on partially observed information. A policy may recommend
rare actions precisely where the data provide the least support. Confounding,
selection, censoring, and changing practice can then be mistaken for treatment
effects. Coverage diagnostics and causal assumptions must precede policy
optimization, not appear as limitations afterward \citep{oberst2019counterfactual,
gottesman2019guidelines}. Cross-site and temporal splits are particularly
important because random splits conceal institutional and policy drift.

\paragraph{3. Estimation: quantify value and uncertainty.}
Off-policy evaluation (OPE) estimates a candidate using data collected under
another policy. Importance sampling, model-based estimators, and doubly robust
methods trade bias, variance, and reliance on modeling assumptions
\citep{jiang2016doubly,thomas2015high}. Balanced estimators provide another route
to stable evaluation under policy mismatch \citep{elmachtoub2023balanced}.
Healthcare studies should report several estimators, effective sample size,
uncertainty intervals, and simple clinical baselines. Offline algorithms such as
conservative Q-learning and implicit Q-learning reduce extrapolation beyond the
data but cannot repair unidentified causal effects
\citep{kumar2020conservative,kostrikov2022offline}. Algorithmic conservatism is
therefore evidence about statistical behavior, not a certificate of clinical
safety.

\paragraph{4. Stress testing: search for plausible failure.}
Average estimated value can hide harm to smaller groups, sensitivity to reward
weights, or collapse under shifted practice. Stress tests should vary outcome
definitions, unmeasured-confounding assumptions, transition dynamics, resource
levels, and subgroup composition. Long-run fairness is especially relevant when
today's allocation changes tomorrow's state \citep{jabbari2017fairness,
verma2024group}. Work on no-gain regimes also cautions that more labels or
queries do not automatically resolve uncertainty \citep{kpotufe2022nuances,
yuan2024regimes}. A credible result states the region in which a policy remains
acceptable, not merely the environment in which it wins.

\paragraph{5. Prospective evaluation: test the intervention, not just the policy.}
Retrospective success does not capture alert burden, clinician adaptation,
implementation error, or changes caused by the policy itself. Evidence should
progress through replay and simulation, external validation, silent deployment,
limited decision support, and controlled prospective evaluation as risk permits.
Micro-randomized trials offer a rigorous design for just-in-time adaptive
interventions \citep{nahum2018microrandomized}; conventional randomized trials
remain important when outcomes and risks demand them. Human oversight is not a
binary switch: studies should specify when clinicians can override a policy,
how overrides are analyzed, and whether recommendations alter documentation or
future training data.

\paragraph{6. Lifecycle evidence: maintain validity after launch.}
Deployment changes the data-generating process. Patient populations, workflows,
capacity, and clinical standards evolve; users may also learn to anticipate the
system. Monitoring must therefore include action distributions, outcome and
subgroup metrics, uncertainty, overrides, and changes in data quality. Material
drift should trigger review, rollback, or prospective re-evaluation. Foundation
models and language interfaces may broaden access to sequential optimization
\citep{zhao2025foundation,behari2024decision}. Meta-level agent systems further
connect reinforcement learning, healthcare workflows, and coordinated AI for
social impact \citep{zhao2025maai}. These systems also add translation and
coordination layers whose objectives and recommendations require separate
validation.

\section{What Transfers Across Settings---and What Does Not}

Several principles transfer. First, a policy comparison is only as credible as
its support and causal assumptions. Second, uncertainty should govern the scope
of deployment. Third, evaluation must follow outcomes across time and groups.
Fourth, the intervention includes its interface, users, and operational
constraints. These principles connect treatment learning, engagement systems,
and resource allocation even when their algorithms differ.

Other evidence does not transfer automatically. A treatment policy validated in
one hospital may fail under another documentation or staffing process. A mobile
intervention can permit rapid randomization that would be unethical for a
high-risk treatment. An allocation policy may be individually low risk yet
produce population-level inequity through repeated denial of service. Claims
should therefore be indexed by site, population, workflow, action authority,
and evaluation design. Generalization is a question to be tested, not a default
property of a pretrained policy.

\section{Cross-Domain Patterns in the Evidence}

\subsection{The field is rich in policies but poor in intervention evidence}

Across application areas, the literature is dominated by retrospective policy
development. Acute-care papers commonly learn from one or two intensive-care
databases and compare estimated returns, action concordance, or predicted
outcomes. Mobile-health studies more often include randomization, but frequently
measure a proximal behavioral response rather than a durable clinical endpoint.
Operations studies can evaluate policies in detailed simulators, yet their
transition models may omit organizational responses. Thus, each domain has an
evidentiary strength that another lacks: treatment studies offer clinically
meaningful outcomes, digital interventions offer experimental variation, and
operations research offers explicit resource constraints. A mature field
should combine these strengths rather than treating the domains as separate
literatures.

The evidence ladder clarifies why retrospective performance is an early result,
not a failed version of a clinical trial. Offline studies can identify
promising action patterns, reveal heterogeneity, and rule out policies with poor
support. They can also compare formulation choices before exposing patients or
staff to an intervention. The problem arises when a high estimated return is
described as clinical improvement. OPE estimates a counterfactual under stated
assumptions; it does not measure usability, adherence, workflow adaptation, or
equilibrium effects. Even sophisticated doubly robust estimators share the
limitations of the recorded state and behavior policy
\citep{jiang2016doubly,kallus2020double,uehara2020minimax}.

\subsection{Reward specification is an empirical and normative bottleneck}

Healthcare rewards rarely arrive as a single, timely, uncontested signal.
Mortality is important but sparse; length of stay can reward premature
discharge; adherence can ignore whether treatment is beneficial; and throughput
can conflict with continuity or equity. Composite rewards make optimization
possible by assigning weights, but those weights embed judgments about timing,
severity, and trade-offs. Sensitivity to reward definitions should therefore be
a primary analysis rather than a minor ablation. If small changes produce
qualitatively different policies, the study has discovered a specification
problem, not selected a winner.

Reward learning from preferences or language may make specification more
accessible, but it does not remove the normative choice. Language-conditioned
systems can translate program goals into executable objectives
\citep{behari2024decision}; foundation-model agents may help assemble domain
knowledge and coordinate analytic tasks \citep{zhao2025foundation}. Their
outputs require traceability from stakeholder statement to reward term,
constraint, and downstream behavior. Otherwise, fluency can obscure rather
than resolve disagreement. In high-stakes care, hard clinical exclusions should
not be represented only as large negative rewards, because optimization can
trade them away.

\subsection{External validity is policy dependent}

Predictive generalization asks whether a model remains accurate in a new
population. Policy generalization asks a harder question: whether accuracy is
maintained on the states visited after the new policy changes actions. A policy
that initially transfers well may create a new distribution on which its state
representation and transition model are unreliable. This is especially
important for aggressive policies that depart substantially from historical
practice. Behavior-policy distance is therefore not merely a statistical
quantity; it is an indicator of how much of the proposed intervention remains
untested.

External validation should vary more than hospital identity. Relevant axes
include calendar time, staffing, clinical protocol, data-collection system,
resource capacity, and action availability. Operations policies should be
tested under demand surges and staffing shortages; engagement policies under
habituation and missing responses; and treatment policies under changes in
case mix and standard of care. Robust MDP methods formalize ambiguity in
transitions \citep{iyengar2005robust,nilim2005robust}, but ambiguity sets should
be grounded in observed or clinically plausible change. A mathematical worst
case that excludes the actual deployment shift provides false comfort.

\subsection{Uncertainty must change action}

Many studies quantify uncertainty without specifying what it does. In an
evidence-centered system, uncertainty should determine whether to recommend,
defer, collect information, restrict the action set, or escalate to a human.
This turns calibration into an operational design question. A policy that
abstains appropriately can have lower coverage but greater clinical value than
one that always acts. High-confidence OPE and safe policy improvement provide
formal starting points \citep{thomas2015high,laroche2019safe}, while constrained
and Lyapunov methods address explicit safety costs
\citep{achiam2017constrained,chow2018lyapunov}. Their guarantees remain
conditional on the chosen state, cost, and uncertainty model.

The appropriate response also depends on reversibility. A low-cost message can
support cautious online learning; a medication or device action may require
prior evidence and close oversight; a population allocation may be reversible
for one individual yet accumulate inequity over time. Reporting a single
``safe RL'' label erases these distinctions. Studies should state the failure
being controlled, its time scale, and the actor authorized to intervene.

\section{A Reporting Standard for Healthcare RL Surveys and Studies}

The literature would be easier to compare if each study exposed the same core
evidence. We propose an \emph{evidence card} aligned with the ladder. It is not
a regulatory certification or a new scalar score. It is a structured record of
what was established, under which assumptions, and what remains unknown.

\paragraph{Decision context.}
The card should identify the decision maker, affected population, cadence,
horizon, available actions, and whether recommendations are advisory or
autonomous. It should distinguish a recorded order from a delivered treatment
and state how concurrent clinician actions enter the environment. For
allocation systems, it should specify the shared budget and whether individuals
interact through queues, staff, geography, or network effects. These details
determine whether an MDP, contextual bandit, restless bandit, or another model
is a defensible abstraction.

\paragraph{Objective and constraints.}
Authors should list every reward component, its scale and timing, all hard and
soft constraints, and the stakeholders involved in selecting them. At least
one sensitivity analysis should vary consequential weights or horizons. When a
surrogate outcome is used, the card should explain the evidence linking it to
patient benefit. When multiple outcomes cannot be combined without a contested
trade-off, a Pareto frontier or distributional report is more informative than
one weighted return.

\paragraph{Data provenance and identification.}
The record should describe sites, calendar periods, inclusion criteria,
missingness, censoring, action frequency, and changes in clinical practice.
Sequential positivity should be assessed for clinically important states, not
only summarized globally. The causal graph or equivalent assumptions should
identify time-varying confounders and variables that are unavailable at the
decision time. Dataset size alone does not establish support: millions of
records can contain almost no evidence for a rare action in a rare state.

\paragraph{Policy learning and comparison.}
Reports should separate representation learning, transition or outcome models,
policy optimization, and hyperparameter selection. Baselines should include
observed practice, simple rules, and clinically recognized standards where
possible, not only other deep RL algorithms. All policies compared by OPE
should be evaluated on held-out trajectories that played no role in tuning.
Estimate-then-optimize comparisons are relevant because errors that look small
predictively can induce different decisions \citep{elmachtoub2023estimate}.

\paragraph{Evaluation and uncertainty.}
At least two OPE families should be reported when feasible, together with
confidence intervals, effective sample size, behavior-policy distance, and the
fraction of decisions outside reliable support. Model-based simulation should
include validation of transition and outcome predictions along policy-relevant
trajectories. Disagreement across estimators is itself a result. It should not
be hidden by choosing the estimate most favorable to the proposed policy.

\paragraph{Robustness, fairness, and failure analysis.}
The card should report performance by clinically meaningful group, site, and
time period with uncertainty. Stress tests should cover reward changes,
unmeasured confounding, data corruption, capacity, and plausible transition
shift. Authors should document policies rejected during development and the
conditions under which the selected policy becomes unacceptable. This makes
negative evidence reusable and reduces pressure to present every learned policy
as deployment ready.

\paragraph{Prospective status and governance.}
Finally, the card should name the highest rung actually reached: retrospective
analysis, external validation, silent deployment, assisted use, randomized
evaluation, or monitored operation. It should describe overrides, incident
response, rollback, re-evaluation triggers, and responsibility for approval.
Model cards and lifecycle documentation offer useful precedents, but healthcare
RL needs policy-specific information about action authority, feedback, and
long-run consequences. Clear status labels would prevent a simulation result
from being mistaken for clinical validation.

\section{Implications for Future Reviews}

Future surveys should move beyond counting papers by algorithm. A useful review
would code each study by application family, data source, causal assumptions,
policy distance, OPE methods, external validation, prospective status, and
governance. This would reveal where evidence is accumulating and where many
papers repeatedly occupy the same early rung. It would also support narrower
systematic reviews: for example, whether sepsis policies have external support,
whether mobile-health effects persist beyond proximal rewards, or whether
allocation studies report distributional consequences.

The evidence ladder can also organize research funding. Early-stage grants may
appropriately support formulation and retrospective feasibility; translational
programs should require cross-site validation, workflow design, and prospective
protocols; deployment funding should include monitoring infrastructure and
institutional ownership. Requiring every paper to claim immediate clinical
impact encourages overstatement. Recognizing distinct rungs allows rigorous
work at each stage while preserving a clear path toward intervention.

\section{Gaps and Research Agenda}

\paragraph{Benchmark evidence, not only return.}
Shared datasets should include behavior-policy information, temporal and site
splits, missingness mechanisms, subgroup labels, and clinically defensible
baselines. Leaderboards should separate identification, estimation, robustness,
and prospective evidence rather than collapse them into one return estimate.

\paragraph{Connect decision quality to workflow outcomes.}
Research should evaluate whether recommendations are timely, interpretable,
actionable, and compatible with team-based care. Cognitive models of engagement
and sequential representations can help characterize how users respond over
time \citep{seow2024improving}, but evaluation must include burden, overrides,
and unintended adaptations.

\paragraph{Make negative evidence cumulative.}
The field learns little when unsupported actions, estimator disagreement, and
failed rollouts remain unpublished. Standardized evidence cards could record
the ladder rung reached, assumptions tested, failure boundaries, and reasons a
policy was not advanced. This would reward useful falsification alongside
algorithmic improvement.

\section{Conclusion}

RL is well matched to the sequential structure of healthcare, but the path from
historical data to beneficial intervention is longer than an optimization
pipeline. Our evidence ladder complements existing algorithm- and
application-centered surveys by organizing the field around what each stage can
justify. The most important frontier is not simply a higher estimated return.
It is a cumulative chain of evidence connecting a clinically meaningful
problem, identifiable comparisons, reliable evaluation, robust behavior,
prospective benefit, and accountable monitoring.

\clearpage
\bibliographystyle{plainnat}
\bibliography{healthcare_rl_evidence_ladder}
\end{document}